%% file: main.tex
\documentclass[10pt,twocolumn,letterpaper]{article}
\usepackage{style}
\definecolor{colorblue}{rgb}{0.21,0.49,0.74}
\usepackage[pagebackref,breaklinks,colorlinks,allcolors=colorblue]{hyperref}
\usepackage{cuted}
\usepackage{multirow}
\usepackage{booktabs}
\def\paperID{0000}
\def\confName{NONE}
\def\confYear{2025}
\newcommand{\blfootnote}[1]{\begingroup
	\renewcommand\thefootnote{}\footnote{#1}\addtocounter{footnote}{-1}
	\endgroup}
\title{ToonComposer: Streamlining Cartoon Production with \\ Generative Post-Keyframing}
\author{
Lingen Li$^{1,2}$ \quad Guangzhi Wang$^{2}$$^\star$$^\dagger$ \quad Zhaoyang Zhang$^{2}$$^\star$$^\dagger$ \quad Yaowei Li$^{3,2}$ \quad Xiaoyu Li$^{2}$ \quad \\Qi Dou$^{1}$ \quad Jinwei Gu$^{1}$ \quad Tianfan Xue$^{1}$$^\dagger$ \quad Ying Shan$^{2}$\\
{
$^{1}$The Chinese University of Hong Kong \
$^{2}$ARC Lab, Tencent PCG \
$^{3}$Peking University}
}
\begin{document}
\maketitle

\blfootnote{$\star$ Project Lead.}
\blfootnote{$\dagger$ Corresponding authors.}
\begin{strip}
\vspace{-70pt}
\centering
\includegraphics[width=1.0\textwidth]{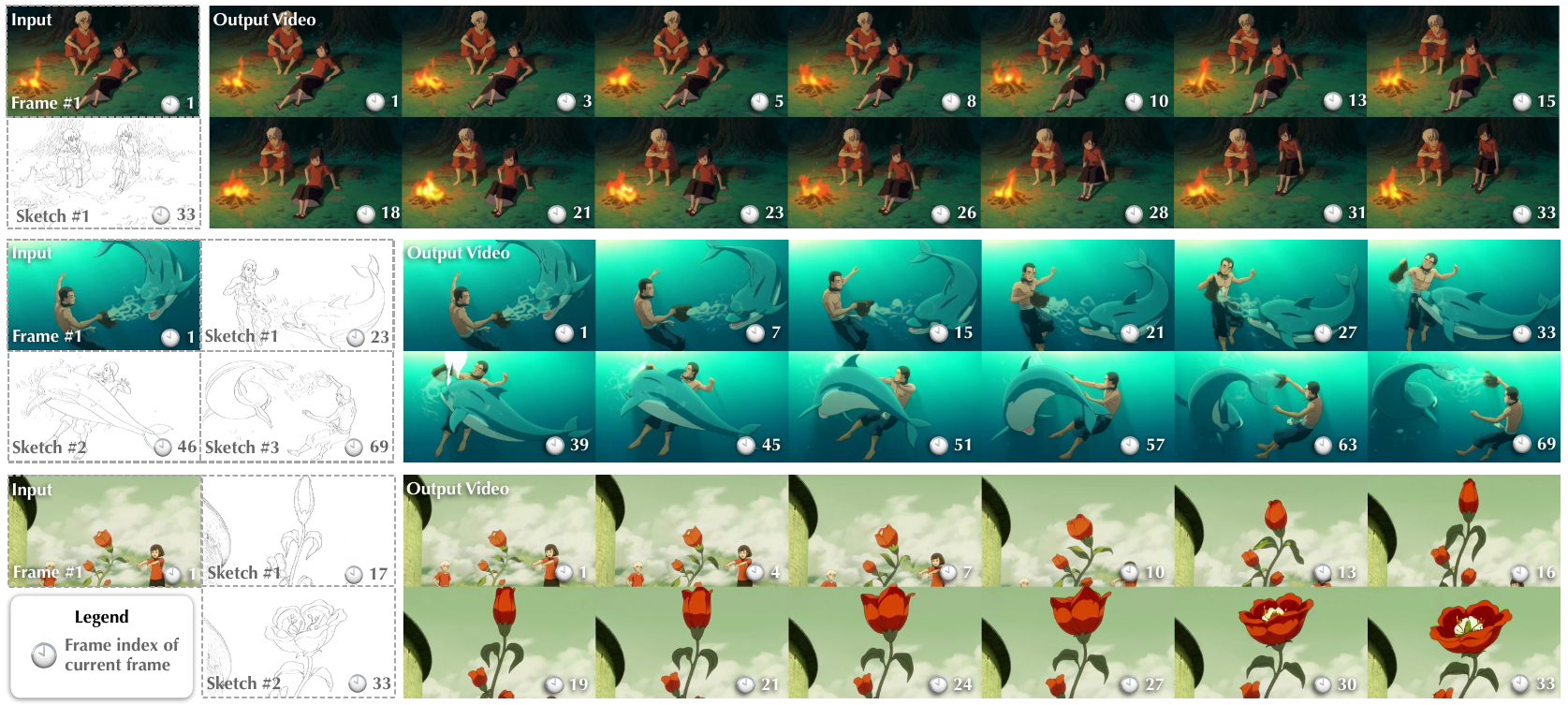}
\vspace{-20pt}
\captionof{figure}{
Video samples generated by ToonComposer using sparse keyframe sketches, featuring scenes from cartoon movies (\textit{Big Fish \& Begonia}, used with permission). ToonComposer supports precise keyframe control and flexible inference with varying numbers of input keyframe sketches and output video lengths (33 frames for the first and third samples, and 69 frames for the second sample). The frames are evenly sampled for illustration in this figure. Each input and output frame is annotated with its corresponding temporal index in the bottom right corner. These movies were excluded from the training data.
}
\label{fig:teaser}
\end{strip}

\begin{abstract}
Traditional cartoon and anime production involves keyframing, inbetweening, and colorization stages, which require intensive manual effort. 
Despite recent advances in AI, existing methods often handle these stages separately, leading to error accumulation and artifacts. For instance, inbetweening approaches struggle with large motions, while colorization methods require dense per-frame sketches. To address this, we introduce ToonComposer, a generative model that unifies inbetweening and colorization into a single post-keyframing stage. ToonComposer employs a sparse sketch injection mechanism to provide precise control using keyframe sketches. Additionally, it uses a cartoon adaptation method with the spatial low-rank adapter to tailor a modern video foundation model to the cartoon domain while keeping its temporal prior intact.
Requiring as few as a single sketch and a colored reference frame, ToonComposer excels with sparse inputs, while also supporting multiple sketches at any temporal location for more precise motion control. This dual capability reduces manual workload and improves flexibility, empowering artists in real-world scenarios. To evaluate our model, we further created PKBench, a benchmark featuring human-drawn sketches that simulate real-world use cases. Our evaluation demonstrates that ToonComposer outperforms existing methods in visual quality, motion consistency, and production efficiency, offering a superior and more flexible solution for AI-assisted cartoon production. Project page: \url{https://lg-li.github.io/project/tooncomposer}.
\end{abstract}
\input{body}
{
    \small
    \bibliographystyle{bib_style}
    \bibliography{main.bib}
}
\end{document}

%% file: body.tex
\input{vars}

\section{Introduction}
\label{sec:intro}
Cartoons and anime are celebrated for their vibrant aesthetics and intricate narratives, standing as a cornerstone of global entertainment.
Traditional cartoon production involves keyframing, inbetweening, and colorization stages, each of which requires artists to craft numerous frames to ensure fluid motion and stylistic consistency~\cite{tang2025generative}. 
While the keyframing stage is a creative process that embodies human artistry, the subsequent inbetweening and colorization stages are highly labor-intensive and time-consuming. Specifically, the inbetweening and colorization stages, which require less creative input, demand the production of hundreds of drawings for mere seconds of animation, resulting in significant time and resource costs.

Recent advances in generative models have facilitated the inbetweening and colorization stages, such as ToonCrafter~\cite{xing2024tooncrafter,jiang2024exploring}, AniDoc~\cite{meng2024anidoc}, or LVCD~\cite{huang2024lvcd}. However, these methods face critical limitations: (1) inbetweening approaches~\cite{xing2024tooncrafter,jiang2024exploring} struggle to interpolate large motions from sparse sketch inputs, often requiring multiple keyframes for a smooth motion; (2) colorization methods~\cite{meng2024anidoc,huang2024lvcd} demand detailed per-frame sketches, imposing significant artist workload. (3) Additionally, their sequential processing leads to error accumulation, where inaccuracies in interpolated sketches affect the colorization stage, resulting in artifacts and reduced quality~\cite{tang2025generative}. These shortcomings highlight a significant gap in achieving a streamlined and efficient production pipeline that produces high-quality results.

\begin{figure}[t]
    \centering
    \includegraphics[width=1\linewidth]{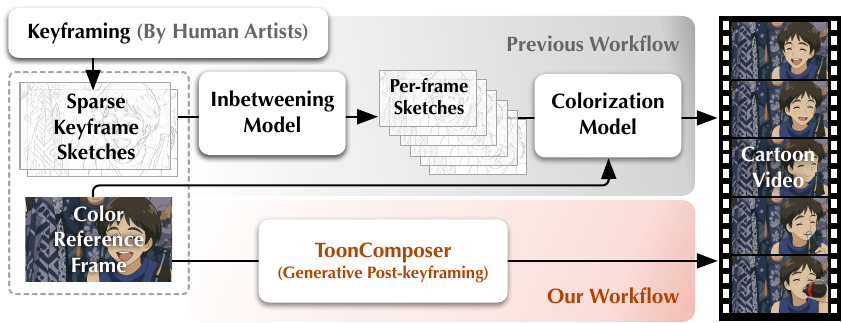}
    \vspace{-7mm}
    \caption{Comparison between previous cartoon production workflow and ours. ToonComposer enables the \textit{\newstagename} stage, seamlessly integrating inbetweening and colorization into a single automated process, streamlining cartoon production compared to previous traditional and existing AI-assisted workflows. }
    \vspace{-4mm}
    \label{fig:workflow-cmp}
\end{figure}

In fact, the inbetweening and colorization stages are highly interdependent. Both of them require searching for correspondences among keyframe sketches or color reference frames. 
Therefore, we introduce the \textbf{\newstagename} stage, a novel paradigm that follows the keyframe creation stage and merges inbetweening and colorization into a single automated process. 
This unification enables the model to jointly utilize the elementary and style information in keyframe sketches and reference frames in a single stage, avoiding the risk of cross-stage error accumulation.
As illustrated in \Cref{fig:workflow-cmp}, the \newstagename stage requires only a few keyframe sketches and a colored reference frame to generate a complete high-quality cartoon video. This approach significantly reduces manual effort, allowing artists to focus on creative keyframe design, while AI manages repetitive tasks.

To achieve this, we adopt modern Diffusion Transformer (DiT)-based foundation models~\cite{wang2025wan}, which demonstrate superior video generation performance. 
Although it offers new possibilities for cartoon video production, it also presents two significant challenges:
1) \textit{Controllability}: DiT foundation models are typically weakly conditioned on text prompts or initial frames, lacking the precision needed to incorporate sparse keyframe sketches for motion guidance at a specific temporal position.
2) \textit{Adaptation}: Since they are trained on natural video datasets, adaptation to the cartoon domain is necessary to produce high-quality cartoon videos.
However, the previous adaptation method for cartoon~\cite{xing2024tooncrafter} is limited to UNet-based models, which alters the models' spatial behavior while preserving temporal prior by only tuning the decoupled spatial layers. 
In our context, the full attention mechanism in the DiT model jointly learns spatial and temporal behaviors, where the previous cartoon adaptation method~\cite{xing2024tooncrafter} is no longer valid.

To address the above challenges, we propose \textbf{ToonComposer}, a generative \newstagename model built on the state-of-the-art DiT video foundation model Wan 2.1~\cite{wang2025wan}.
First, to solve the challenge of precise controllability with sparse sketches, we devise the \textit{sparse sketch injection} mechanism, which enables accurate control in cartoon generation using sparse keyframe sketches.
Second, to handle the challenge of domain adaptation in DiT models, we tailor the \textit{cartoon adaptation} for ToonComposer. 
It adapts the foundation DiT model to the cartoon domain with a novel spatial low-rank adapter (SLRA) strategy, which adapts the appearance to the cartoon domain while preserving its powerful temporal prior intact.
In addition, we improve the flexibility of ToonComposer by introducing the \textit{region-wise control}, which enables flexible motion generation without drawing sketches in indicated regions.
These contributions ensure that ToonComposer generates stylistically coherent animations with minimal input, as shown in \Cref{fig:teaser}, effectively realizing the \newstagename stage within one model.

To support the training of the proposed model, we curated a large-scale dataset \textit{\datasetname}, which contains high-quality anime and cartoon video clips. 
Each clip is accompanied by keyframe sketches in multiple styles, providing a solid foundation for training.
In addition to evaluating our model on synthetic benchmark, we curated \benchmarkname, a new benchmark that contains 30 original cartoon scenes with human-drawn keyframe sketches and reference color frames.
Extensive experiments on both benchmarks demonstrate that ToonComposer outperforms existing methods in visual quality, motion coherence, and production efficiency. 

Our contributions are summarized as follows:
\begin{itemize}
    \item We introduced the \newstagename stage, a new cartoon production paradigm that integrates inbetweening and colorization into a single AI-driven process, significantly reducing manual labor.
    \item We proposed ToonComposer, the first DiT-based cartoon generation model for \newstagename,  incorporating sparse sketch injection and region-wise control to generate high-quality cartoon videos from sparse inputs.
    \item We design a cartoon adaptation mechanism using SLRA, a novel low-rank adaptation strategy that effectively tailors the spatial behavior of the DiT model to the cartoon domain while preserving its temporal prior.
    \item We curate a dataset of cartoon video clips with diverse sketches for training and develop a high-quality benchmark with real human-drawn sketches, PKBench, for cartoon \newstagename evaluation.
\end{itemize}

\begin{figure*}
    \centering
    \includegraphics[width=1\linewidth]{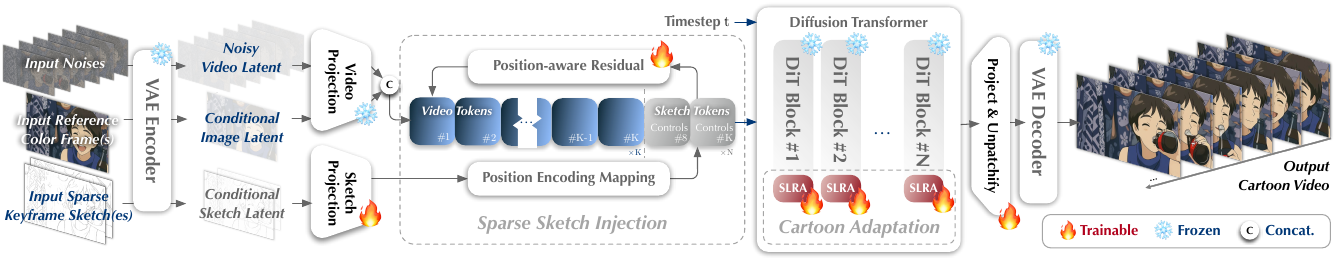}
    \vspace{-5mm}
    \caption{The model design of ToonComposer. A sparse sketch injection mechanism enables precise control using keyframe sketches, and a cartoon adaptation method incorporating a spatial low-rank adapter tailors the DiT-based video foundation model to the cartoon domain, preserving its temporal priors.}
    \label{fig:method}
\end{figure*}

\section{Related Work}
\label{sec:related}
\paragraph{AI-assisted Cartoon Production} AI has increasingly been applied to automate labor-intensive tasks in cartoon and anime production, such as inbetweening and colorization. For inbetweening, early methods like AnimeInterp~\cite{li2021deep} and AutoFI~\cite{shen2022enhanced} focus on linear and simple motion interpolation. More recently, diffusion-based methods~\cite{xing2024tooncrafter,jiang2024exploring} become capable of handling cases with more complex motion by harnessing the generative priors of a pretrained model.
For colorization, early GAN-based~\cite{isola2017image} and recent diffusion-based methods~\cite{zhuang2024colorflow,meng2024anidoc,huang2024lvcd,zhuang2025cobra} have automated the colorization of line art based on one or a series of reference frames.

However, while existing AI-assisted methods have advanced cartoon production by automating inbetweening and colorization, they often require dense frame inputs, operate as separate, isolated stages, and face challenges with complex motions and stylistic consistency~\cite{tang2025generative}. ToonComposer overcomes these hurdles by offering a unified, sparse-input solution for post-keyframing that simplifies the production workflow.

\vspace{-2mm}
\paragraph{Video Diffusion Model}
Diffusion models have emerged as the cornerstone for generative tasks~\cite{ho2020denoising}, particularly in image and video synthesis, by iteratively denoising samples from a noise distribution to produce high-quality outputs~\cite{blattmann2023align}. For video generation, these models must effectively capture both spatial details and temporal dynamics, a challenge that has led to distinct architectural approaches. Traditional UNet-based diffusion models~\cite{ho2022video,blattmann2023align,blattmann2023stable,xing2024dynamicrafter} extend 2D U-Nets to handle videos by incorporating 3D convolutions and separated spatial and temporal attention layers. In these models, spatial attention layers process intra-frame features, often across channels or spatial positions, while separate temporal attention layers model inter-frame dependencies. In contrast, Diffusion Transformers (DiTs)~\cite{peebles2023scalable} leverage transformer architectures, replacing UNet’s convolutional backbone with full attention mechanisms that model long-range dependencies in both spatial and temporal dimensions~\cite{yang2024cogvideox,kong2024hunyuanvideo,wang2025wan}. Although showing stronger performance compared to spatial-temporal decoupled design, such full attention mechanism eliminates the availability of spatial adaptation tailored for domains such as cartoon~\cite{xing2024tooncrafter}.
Our work builds upon the DiT-based foundation model to harness the high-quality video prior with a new cartoon adaptation mechanism, which adapts the DiT-based foundation model to the cartoon domain in spatial behavior while keeping its temporal motion prior intact.

\vspace{-2mm}
\paragraph{Controllable Generation}
Controllable generation seeks to steer image and video synthesis with explicit conditions such as reference images~\cite{ye2023ip,zhang2023adding}, depth maps~\cite{xing2024make}, human poses~\cite{zhu2024champ,hu2024animate}, and semantic labels. Techniques such as IP-Adapter~\cite{ye2023ip} and ControlNet~\cite{zhang2023adding} inject these visual cues into diffusion models alongside text prompts, allowing fine-grained manipulation of both content and style.
The value of controllability of generation is particularly evident in domain-specific pipelines. For cinematography, camera-aware generators expose handles for 2D scene layout and 3D camera trajectories~\cite{he2024cameractrl,wang2024motionctrl,li2024image,wang2025cinemaster}, allowing video creators to frame shots and motion with high precision. In cartoon production, sketch-guided models support interpolation, inbetweening, and colorization~\cite{meng2024anidoc,huang2024lvcd,xing2024tooncrafter}. Our method focuses on the controllable cartoon generation using sparse keyframe sketches to accelerate the production process.

\section{Methodology}
We introduce ToonComposer, a novel generative post-keyframing model that produces high-quality cartoon videos with sparse control.
To achieve this, we propose a curated sparse sketch injection strategy, which effectively enables precise sketch control at arbitrary timestamps (Sec.~\ref{sec:sketch-injection}). 
Furthermore, to fully leverage the temporal prior in video generation models, we design a novel low-rank adaptation strategy that efficiently adapts the spatial prior to the cartoon domain while leaving the temporal prior intact (Sec.~\ref{sec:cartoon-adaptation}).
To further alleviate artist workload and improve efficiency, our method also enables region-wise control, empowering artists to draw only part of the sketches while leaving the model to reason how the motion should be generated in blank areas (Sec.~\ref{sec:region-control}).

\subsection{\newstagenamecaptial Stage}\label{sec:post-keyframing-stage}
In recent years, the cartoon industry has benefited significantly from the development of generative AI, facilitating the stage of inbetweening~\cite{xing2024tooncrafter} and colorization~\cite{huang2024lvcd, meng2024anidoc}. 
Although they are helpful in cartoon video production, existing methods are often bottlenecked by high labor demand or low video quality. 
For example, colorization methods often require one colored reference frame and per-frame sketches, which is expensive to obtain. 
Although recent inbetwenning methods~\cite{xing2024tooncrafter} can be utilized to generate per-frame sketches, it still faces challenges with large motions, leading to error accumulation in the colorization stage.

In fact, the two stages are highly interdependent: both require searching and interpolating along the correspondence between elements in the keyframes/sketches, indicating that their internal mechanisms are similar. 
Inspired by this, we propose the \textit{\newstagename} stage, a new stage that automates cartoon production and consolidates the inbetweening and colorization into a unified generative process.
Given \textbf{one} colored reference frame and \textbf{one} sketch frame, the \textit{\newstagename} stage aims to directly produce a high-quality cartoon video that adheres to the guidance provided by these inputs. 
This process significantly alleviates the requirement of dense per-frame sketches, avoiding the risk of cross-stage error accumulation.

Formally, given a colored reference frame $f_1$ and a sketch frame $s_j$, we aim to obtain a model $\mathbf{G}_\theta$ that directly generates a high-quality cartoon video with $K$ frames:
\begin{equation}
\{\hat{f}_k\}_{k=1}^K = \mathbf{G}_\theta(f_1, s_j, e_{text})
\end{equation}\label{eq:new-stage}
where $j$ represents the temporal location of $s_j$, and $e_{text}$ represents the text prompt describing the scene.

In this work, we adopt the recently proposed powerful video generation model Wan~\cite{wang2025wan} as the basis.

\subsection{Sparse Sketch Injection}
\label{sec:sketch-injection}
Advanced video generation models, such as Wan~\cite{wang2025wan}, demonstrate exceptional performance in producing high-quality videos. While its image-to-video (I2V) variant supports video generation guided by an initial frame, precise control using sparse sketches at arbitrary temporal positions remains unexplored.

To this end, we introduce a novel \textit{sparse sketch injection}mechanism that seamlessly integrates sketches into the latent token sequence of an I2V DiT model for precise temporal control. In a standard I2V DiT model $\epsilon_{\theta}$, the input image is concatenated with the noisy latent $z$ along the channel dimension:
\begin{equation}
    \hat\epsilon = \epsilon_{\theta}\left( [\{z_k^{(t)}\}_{k=1}^{K}, \text{pad}(f'_{1})]_{\text{c}}, e_{\text{text}}, t \right),
    \label{eq:i2v}
\end{equation}
where $\hat\epsilon$ denotes model output, $z_k^{(t)}$ is the noisy latent tokens of the $k$-th frame at timestep $t$, $\text{pad}(\cdot)$ represents the zero padding along the temporal dimension, $[,]_c$ represents concatenation along the channel dimension, and $f'_1$ represents the latent of $f_1$ encoded by the VAE.
Our sparse sketch injection mechanism enhances the model $\epsilon_{\theta}$ to support precise control using sparse sketches and keyframes through a position encoding mapping and a position-aware residual mechanism.

\paragraph{Position Encoding Mapping}
To inject the sketch frame $ s_j$ into the latent representation of the DiT model $\epsilon_{\theta}$ for precise control over the temporal location $j$ in the generated cartoon, 
we first introduce an additional projection head that embeds the conditional sketch latents into sketch tokens $ s'_j$ that are compatible with the latent dimension of the model. 
Then, we apply the \textit{position embedding mapping} process that borrows the RoPE~\cite{su2024roformer} encodings from the corresponding video tokens at the temporal index $j$ before each attention operator.
These sparse sketch tokens are concatenated with the video tokens along the sequence dimension to facilitate the attention process.

This mechanism enables efficient integration of sketch conditions into the latent space with temporal awareness during the generation process. In addition, it facilitates the simultaneous use of multiple keyframes and sketches as control input. Given the complexity of motion in some cartoon scenes, precise control often necessitates multiple keyframes and sketches. Therefore, we extend the formulation to support both multiple colored reference frames and multiple sketch inputs. Consequently, the forward step of the DiT model is expressed as:
\begin{equation}
    \hat\epsilon = \epsilon_{\theta}\left( \left[[\{z_k^{(t)}\}_{k=1}^{K}, \text{pad}(\{f'_{i_c}\}_{c=1}^{C})]_{\text{c}}, \{ s'_{i_n} \}_{n=1}^{N}\right]_{\text{s}}, e_{\text{text}}, t \right),
    \label{eq:sparse-sketch-inj}
\end{equation}
where $\{ s'_{i_n} \}_{n=1}^{N}$ represents $N$ sketch frames and $\{f'_{i_c}\}_{c=1}^{C})$ denotes $C$ colored reference frames.
$[\cdot, \cdot]_{\text{s}}$ means concatenation along the token sequence dimension. This formulation enables precise control over multiple inputs, while also supporting the minimal input requirement of the \textit{\newstagename} stage (one colored and one sketch frame, described in \Cref{eq:new-stage}).

\paragraph{Position-aware Residual}
To enhance the flexibility of sketch control, our ToonComposer also allows users to dynamically adjust the control strength of input sketches, through an adjustable weight during inference. This is done through a new \textit{positional-aware residual} module in the injection process. 
For sparse sketch tokens at controlled keyframe indices $\{i_n\}_{n=1}^N$, we transform these tokens via a linear layer $W_{res}$ and add them to the corresponding video tokens with the same indices with a scaling weight $\alpha$:
\begin{equation}
    \{z_k^{(t)}\}_{k\in\{i_n\}_{n=1}^N} := \{z_k^{(t)}\}_{k\in\{i_n\}_{n=1}^N} + \alpha \{ s'_{i_n} \}_{n=1}^N W_{res},
    \label{eq:pos-aware-residual}
\end{equation}
where $W_{res} \in \mathbb{R}^{D \times D}$ is a trainable weight, $D$ is the feature dimension of tokens. During training, the scaling weight $\alpha$ is set to $1$. During inference, users can adjust the weight $\alpha$ to loosen or strengthen the control of keyframe sketches.

\begin{figure}
    \centering
    \includegraphics[width=0.78\linewidth]{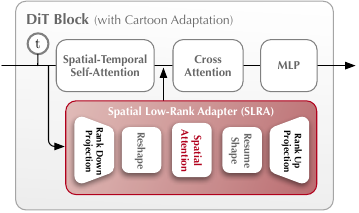}
    \caption{The structure of the Spatial Low-Rank Adapter (SLRA) used for cartoon adaptation in ToonComposer. The SLRA takes the hidden states before the spatial-temporal self-attention module as input and outputs a residual that is added after the self-attention operation.}
    \label{fig:method-slra}
\end{figure}

\subsection{Cartoon Adaptation}
\label{sec:cartoon-adaptation}
Previous work~\cite{xing2024tooncrafter} has demonstrated the success of adapting the video generation model to the cartoon domain. 
By tuning only the spatial layers of the spatial-temporal U-Net, the temporal motion prior in the original model is preserved, while the appearance part is adapted to the cartoon.
However, with the development of video generation models, 3D-full attention has been extensively used in modern video generation models~\cite{yang2024cogvideox, wang2025wan, kong2024hunyuanvideo}, where spatial and temporal representations are intertwined in the latent space.
As a result, it is not feasible to directly perform spatial adaptation as is done in previous work.

To address this, we introduce the Spatial Low-rank Adapter (SLRA), a novel low-rank adaptation mechanism designed to effectively adapt video generation models to the cartoon domain. 
Similar to traditional LoRA~\cite{hu2022lora}, our SLRA also contains two trainable matrices $\mathbf{W}^{\text{down}}$ and $\mathbf{W}^{\text{up}}$. Different from traditional LoRA, our SLRA alters attention module's spatial behavior only, so as to preserve the temporal prior in the base model. 

Let $H$ and $W$ be the spatial sizes of DiT's latent tokens, and $K$ and $N$ be the temporal length of the video tokens and sketch tokens.
Given a token sequence $h \in \mathbb{R}^{L \times D}$ inside each self-attention module of $\epsilon_{\theta}$, where $L = (K + N) \times H \times W$ is the full token sequence length, SLRA operates first by downsampling the feature dimension of the input hidden token with a linear layer: \begin{equation}
    h^{\text{low}} = h \mathbf{W}^{\text{down}},
\end{equation}
where $\mathbf{W}^{\text{down}} \in \mathbb{R}^{D \times D'}$ $D' \ll D$, yielding $h^{\text{low}} \in \mathbb{R}^{L \times D'}$.

Then, SLRA reshapes $h^{\text{low}}$ to $\tilde{h}^{\text{low}} \in \mathbb{R}^{[K + N] \times [H \times W] \times D'}$, recovering their original spatial-temporal arrangements. 
After reshaping, we perform a self-attention operation on spatial dimension only. This is achieved by performing the attention mechanism on the spatial dimension of each frame \textbf{independently}:
\begin{equation}
    Q = \tilde{h}^{\text{low}}_l \mathbf{W}_Q, \quad K = \tilde{h}^{\text{low}}_l \mathbf{W}_K, \quad V = \tilde{h}^{\text{low}}_l \mathbf{W}_V
\end{equation}
\begin{equation}
    O = \text{softmax}\left( \frac{Q K^T}{\sqrt{D'}} \right) V,
\end{equation}
where the subscript $l$ represents the index along the $l$-th temporal dimension. Thus, the attention computation is performed within each frame.
$\mathbf{W}_Q, \mathbf{W}_K$, $\mathbf{W}_V \in \mathbb{R}^{D' \times D'}$ are trainable matrices and $O \in \mathbb{R}^{H \times W \times D'}$. The same positional embeddings as the main model will be applied to both video and sketch tokens during this attention operation. 
As a result, the information propagation in the SLRA module is performed only in the spatial dimension, while leaving the temporal dimension intact.

Following that, we reshape $O$ to $\hat{h}^{\text{low}} \in \mathbb{R}^{L \times D'}$ as a sequence, then upsample it to the original dimension:
\begin{equation}
    h_{\text{res}} = \hat{h}^{\text{low}} \mathbf{W}^{\text{up}},
\end{equation}
where $\mathbf{W}^{\text{up}} \in \mathbb{R}^{D' \times D}$.
Finally, the adapted self-attention output will be modified as: 
\begin{equation}
    h' = \text{SelfAttention}(h) + h_{\text{res}}.
\end{equation}
The operation process of SLRA is illustrated in~\Cref{fig:method-slra}.

SLRA ensures that cartoon-specific spatial features are learned without disrupting temporal coherence, efficiently adapting a DiT-based video diffusion model to the cartoon domain.

\subsection{Region-wise Control}
\label{sec:region-control}
Sometimes cartoon creators may only want to draw the foreground sketch and let the generator create the background for them.
If they simply leave the background blank, this may result in undesirable artifacts, as shown in the second row of~\Cref{fig:exp-case-region-wise-control}.

To this end, we propose a novel region-wise control mechanism that allows artists to specify blank regions in sketches for the model to generate plausible content based on context or text prompts. During training, random masks $m_{i_n} \in \{0,1\}^{H \times W}$ are applied to the sketch frames $s_{i_n}$, where $m_{i_n}(i,j) = 0$ indicates a region where the sketch is not provided. 
An additional channel is concatenated to $s_{i_n}$, which is encoded as:
\begin{equation}
    \tilde{s}'_{i_n} = \left[ \mathcal{E}(s_{i_n}),m_{i_n}\right]_\text{c} ,
\end{equation}
where $\tilde{s}'_{i_n}$ is used to replace the $s'_{i_n}$ described in \Cref{eq:sparse-sketch-inj} during training. The model learns to reconstruct full frames in masked regions, enabling flexible inference where artists can assign the value of $m_{i_n}$ and leave masked areas blank for context-driven generation.

Complementary to the support of temporally sparse keyframes and sketches, our region-wise control allows the input sketch to be spatially sparse, further alleviating the requirements and labors for cartoon creators.

\subsection{Training Objective}
ToonComposer is trained as a conditional diffusion model following Rectified Flow~\cite{esser2024scaling}, which predicts the velocity $v_t$ at a timestep $t$ sampled from logit-normal distribution. 
For simplicity, we write the input part in \Cref{eq:sparse-sketch-inj} as $x_{\text{in}}$:
\begin{equation}
    x_{\text{in}} = \left[[\{z_k^{(t)}\}_{k=1}^K, \text{pad}(\{f'_{i_c}\}_{c=1}^C)]_{\text{c}}, \{ \tilde{s}'_{i_n} \}_{n=1}^N\right]_{\text{s}},
\end{equation}
and let $\mathbf{z}_0 = \{ z_k^{(0)} \}_{k=1}^K$ be a clean cartoon video latent, the training objective minimizes the expected velocity prediction error:
\begin{equation}
    \mathcal{L} = \mathbb{E}_{\mathbf{z}_0, \eta, t} \left[ \left\| v_t - \epsilon_{\theta}\left( x_{\text{in}}, e_{\text{text}}, t \right) \right\|_2^2 \right],
\end{equation}
where $\eta$ is the random Gaussian noise, $v_t$ is the velocity derived from $\{z_k^{(t)}\}_{k=1}^K - \eta$,  and $\epsilon_\theta$ is the ToonComposer model to be trained.

\begin{table*}[t]
\caption{Quantitative evaluation results on the synthetic benchmark, comparing ToonComposer with  previous AI-assisted cartoon generation methods: AniDoc~\cite{meng2024anidoc}, LVCD~\cite{huang2024lvcd}, and ToonCrafter~\cite{xing2024tooncrafter}.}
\resizebox{1\linewidth}{!}{
\begin{tabular}{r|ccc|cccc}
\toprule
Method & LPIPS↓ & DISTS↓ & CLIP↑ & Subject Con.↑ & Motion Smo.↑ & Background Con.↑ & Aesthetic Qua.↑ \\
\midrule
AniDoc~\cite{meng2024anidoc} & 0.3734 & 0.5461 & 0.8665 & 0.9067 & 0.9798 & 0.9408 & 0.4962 \\
LVCD~\cite{huang2024lvcd} & 0.3910 & 0.5505 & 0.8428 & 0.8316 & 0.9810 & 0.9183 & 0.4984 \\
ToonCrafter ~\cite{xing2024tooncrafter} & 0.3830 & 0.5571 & 0.8463 & 0.8075 & 0.9550 & 0.8920 & 0.5035 \\
ToonComposer (Ours) & \textbf{0.1785} & \textbf{0.0926} & \textbf{0.9449}  & \textbf{0.9451} &\textbf{0.9886} & \textbf{0.9547} & \textbf{0.5999} \\
\bottomrule
\end{tabular}
}
\label{tab:eval-synthetic}
\end{table*}

\begin{figure}
    \centering
    \includegraphics[width=1\linewidth]{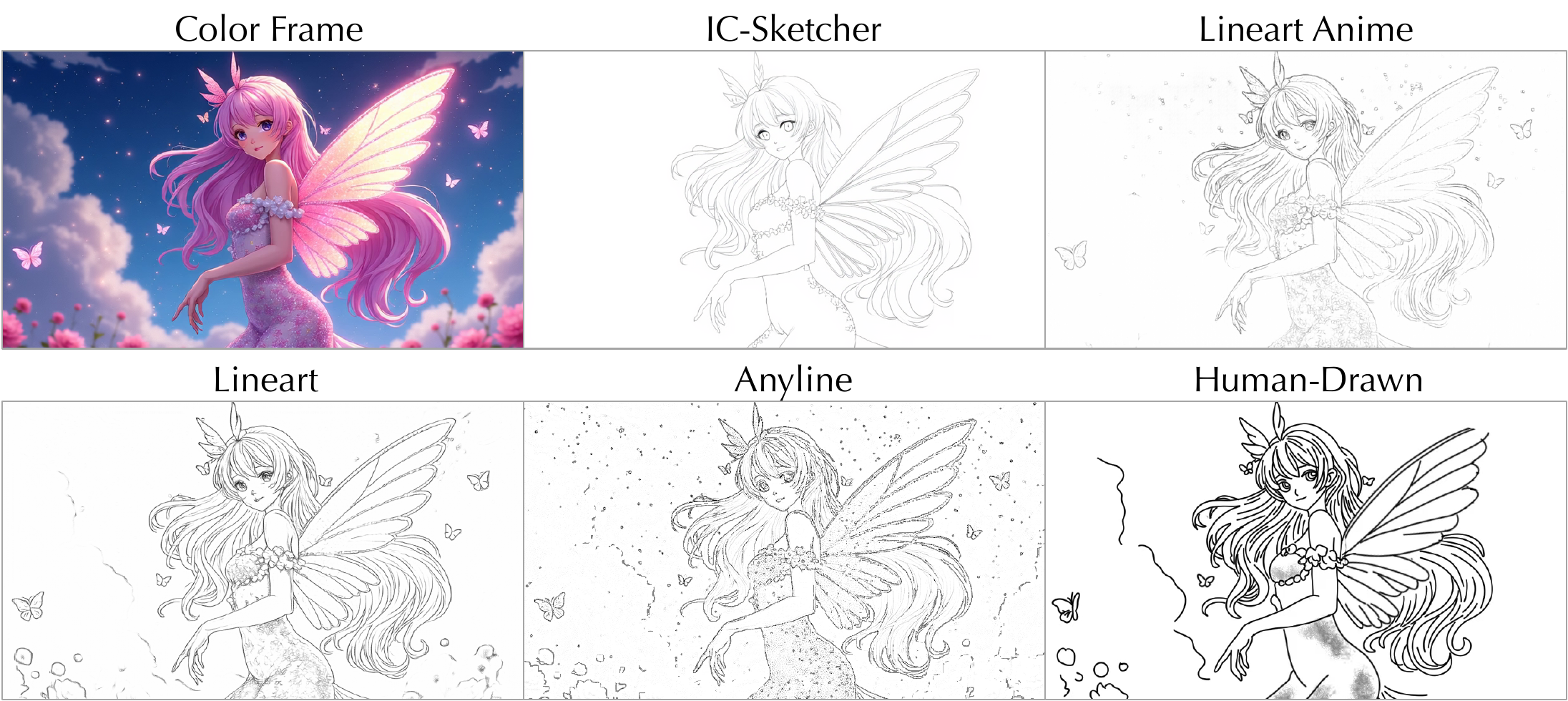}
    \caption{Examples of different sketch types used during training and evaluation. All variants except human-drawn sketches are included in the training set. The diversity of training sketches improves ToonComposer’s robustness to varying sketch styles in real-world use cases. Human-drawn sketches are reserved for evaluation in the real benchmark, as discussed in \Cref{sec:real-bench}.}
    \label{fig:sketch-samples}
\end{figure}

\begin{table}[t]
\caption{Quantitative evaluation results on the real sketch benchmark PKBench, comparing ToonComposer with prior AI-assisted cartoon generation models: AniDoc~\cite{meng2024anidoc}, LVCD~\cite{huang2024lvcd}, and ToonCrafter~\cite{xing2024tooncrafter}.}
\centering
\resizebox{1\linewidth}{!}{
\begin{tabular}{r|cccc}
\toprule
Method & S. C.↑ & M. S.↑ & B. C.↑ & A. Q.↑ \\
\midrule
AniDoc~\cite{meng2024anidoc} & 0.9456 & 0.9842 & 0.9664 & 0.6611 \\
LVCD~\cite{huang2024lvcd} & 0.8653 & 0.9724 & 0.9394 & 0.6479 \\
ToonCrafter~\cite{xing2024tooncrafter}& 0.8567 & 0.9674 & 0.9343 & 0.6822 \\
ToonComposer (Ours) & \textbf{0.9509} & \textbf{0.9910} & \textbf{0.9681} & \textbf{0.7345} \\
\bottomrule
\end{tabular}
}
\label{tab:eval-real}
\end{table}

\section{Experiments}

\subsection{Experimental Settings}

\paragraph{Dataset} 
Based on our internal video sources, we constructed the \datasetname, a high-quality cartoon dataset containing 37K diverse cartoon video clips.
Each clip was accompanied by a descriptive caption generated by CogVLM~\cite{wang2024cogvlm} and a set of sketch frames.
Recognizing the stylistic diversity in sketches due to different artist preferences or creation tools, we augmented our dataset with diverse types of sketches.
Specifically, we synthesize four versions of sketches per frame using four open-source CNN-based sketch models, including two basic lineart models used in ControlNet~\cite{zhang2023adding}, Anime2Sketch~\cite{xiang2022adversarial}, and Anyline~\cite{soria2023tiny}.
Furthermore, we tune a FLUX-based image-to-image generative model with in-context LoRA~\cite{huang2024context} on a small real-sketch dataset from multiple artists. 
This model, named IC-Sketcher, is then used to produce another version of sketches. \Cref{fig:sketch-samples} illustrates one example frame with diverse sketches. 

\paragraph{Benchmark}
We first evaluate our methods on a synthetic benchmark obtained from cartoon movies (use with permission, for evaluation only), where sketches for each video frame are produced using sketch models.
We adopt reference-based metrics on this benchmark since the ground truth is available.
Furthermore, we developed \textit{\benchmarkname}, a novel benchmark featuring human-drawn sketches to enable a more comprehensive evaluation of cartoon \newstagename in real-world scenarios. 
\benchmarkname contains 30 samples, each including 1) a colored reference frame, 2) a textual prompt that describes the scene, and 3) two real sketches that depict the start and end frames of a scene, drawn by professional artists.

\paragraph{Metrics}
For evaluation metrics, we adopt 1) reference-based perceptual metrics for synthetic benchmark, including LPIPS~\cite{zhang2018unreasonable}, DISTS~\cite{ding2020image}, and CLIP~\cite{radford2021learning} image similarity, 
2) reference-free video quality metrics from VBench~\cite{huang2024vbench} for both synthetic and real benchmarks, including subject consistency (S. C.), motion consistency (M. C.), background consistency (B. C.) and aesthetic quality (A. Q.).
3) A user study on human perceptual quality for the real benchmark.

\paragraph{Training Details}
We train our model for 10 epochs using our dataset, with an equivalent batch size of 16, the AdamW~\cite{loshchilov2017decoupled} optimizer, and a learning rate of $10^{-5}$. The rank of SLRA $D_{low}$ is set to 144. We use the zero redundancy optimizer~\cite{rajbhandari2020zero} stage 2 to reduce memory cost during training.

\subsection{Evaluation on Synthetic Benchmark}
We first evaluate our ToonComposer on a synthetic cartoon benchmark and compare it with previous methods, including AniDoc~\cite{meng2024anidoc}, LVCD~\cite{huang2024lvcd}, and ToonCrafter~\cite{xing2024tooncrafter}. 
In this synthetic evaluation, sketches are obtained from cartoon video frames using the same sketch model~\cite{xiang2022adversarial}.
To ensure evaluation fairness, we align the ground truths in both spatial and temporal dimension to fit the pre-defined settings of each model for metrics calculation.

\paragraph{Baseline Methods}
Although our model requires only one inference to get the final cartoon video, previous methods demand a two-stage process, as shown in \Cref{fig:workflow-cmp}. For ToonCrafter~\cite{xing2024tooncrafter}, we first generate the dense sketch sequence by interpolating the first and last sketch frames, then we use its sketch guidance mode (which requires the first and last color frames as input) to generate the final cartoon video. 
For LVCD~\cite{huang2024lvcd} and AniDoc~\cite{meng2024anidoc}, we first generate the dense sketch sequence interpolated by ToonCrafter, then colorize the sketches into a final cartoon video using the two models, respectively.

\paragraph{Results}
\Cref{tab:eval-synthetic} shows the numeric results of the synthetic evaluation. Our method outperforms previous methods in both reference-based metrics and reference-free metrics. For example, our model reports a significantly lower DISTS score, indicating that its perceptual quality is much better than that of its counterparts.

\begin{figure*}[t]
    \centering
    \includegraphics[width=1\linewidth]{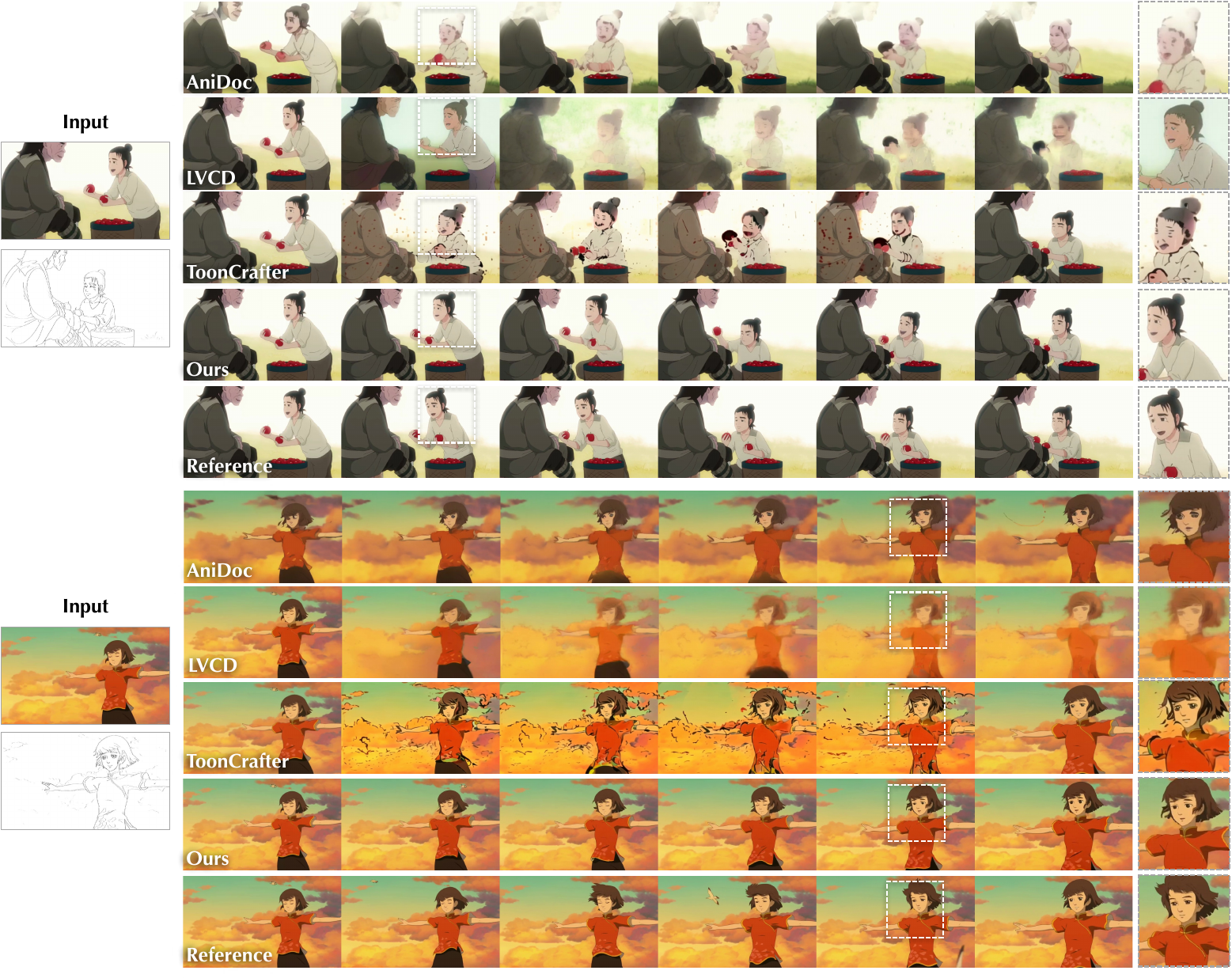}
    \caption{Comparison on the synthetic benchmark among AniDoc, LVCD, ToonCrafter, and our ToonComposer. Zoom-in patches of a randomly selected region are shown in the rightmost column. Our method demonstrates superior visual quality, smoother motion, and better style consistency with the input image. Evaluation scenes are sourced from movies with permission (\textit{Mr. Miao} and \textit{Big Fish \& Begonia}). Please refer to the supplementary video for additional results.}
    \label{fig:exp-comp-synthetic}
\end{figure*}

\Cref{fig:exp-comp-synthetic} visualized the qualitative comparison between these methods, with the ground truth video provided as references.
In both samples, our method produces smooth and natural cartoon video frames, while other methods fail to handle such challenging cases with sparse sketches.
For example, in the zoom-in patches of the first sample, AniDoc and ToonCrafter produce distorted faces. LVCD generates a reasonable face but loses all details in subsequent frames. In contrast, our method produces a clear face which preserves the identity of the first reference frame. These observations align with our method's numeric performance advantages in \Cref{tab:eval-synthetic}. More results are provided in the supplementary video.

\begin{table}[t]
\caption{User preference rates for aesthetic quality and motion quality of cartoons generated by ToonCrafter~\cite{xing2024tooncrafter}, AniDoc~\cite{meng2024anidoc}, LVCD~\cite{huang2024lvcd}, and ToonComposer.}\label{tab:human_eval}
\centering
\begin{tabular}{r|cc}
\toprule
Method & Aesthetic Q.↑ & Motion Q.↑ \\
\midrule
AniDoc~\cite{meng2024anidoc}                & 4.45\%           & 5.34\%\\
LVCD~\cite{huang2024lvcd}                   & 7.54\%           & 7.91\% \\
ToonCrafter~\cite{xing2024tooncrafter}      & 17.02\%          & 18.19\% \\
ToonComposer (Ours)                         & \textbf{70.99\%} & \textbf{68.58\%}  \\
\bottomrule
\end{tabular}
\label{tab:user-study}
\end{table}

\begin{table}[t]
\caption{Ablation study on the Spatial Low-Rank Adapter (SLRA) for cartoon adaptation by modifying its internal attention module: temporal-only (Temp. Adapt.), spatial-temporal (S. T. Adapt.), and removal of the attention module (Linear Adapt.). We also compare against a baseline model adapted using LoRA~\cite{hu2022lora}.}
\centering
\begin{tabular}{ccccccc}
\toprule
Adaptation Method &  LPIPS↓ & DISTS↓ & CLIP↑\\
\midrule
Temp. Adapt.  & 0.1956 & 0.1109 & 0.9581  \\
S. T. Adapt. & 0.1977 & 0.1068 &  0.9587 \\
Linear Adapt. & 0.2030 & 0.1091 & 0.9589  \\
LoRA & 0.1922 & 0.1082 & 0.9628 \\ 
SLRA (Ours) & \textbf{0.1874} & \textbf{0.0955} & \textbf{0.9634} \\
\bottomrule
\end{tabular}
\label{tab:exp-ablation-slra}
\end{table}

\subsection{Evaluation on Real Benchmark}
\label{sec:real-bench}
In addition to the evaluation on the synthetic test set, we further compared all methods on our proposed benchmark \benchmarkname with real human sketches.
Since ground truth is not available for each sample, we evaluated the generated videos using reference-free metrics from VBench~\cite{huang2024vbench}. 
The quantitative comparison is shown in \Cref{tab:eval-real}, where our model outperforms previous methods in all metrics, achieving superior appearance and motion quality.

\begin{figure*}[ht]
    \centering
    \includegraphics[width=0.930\linewidth]{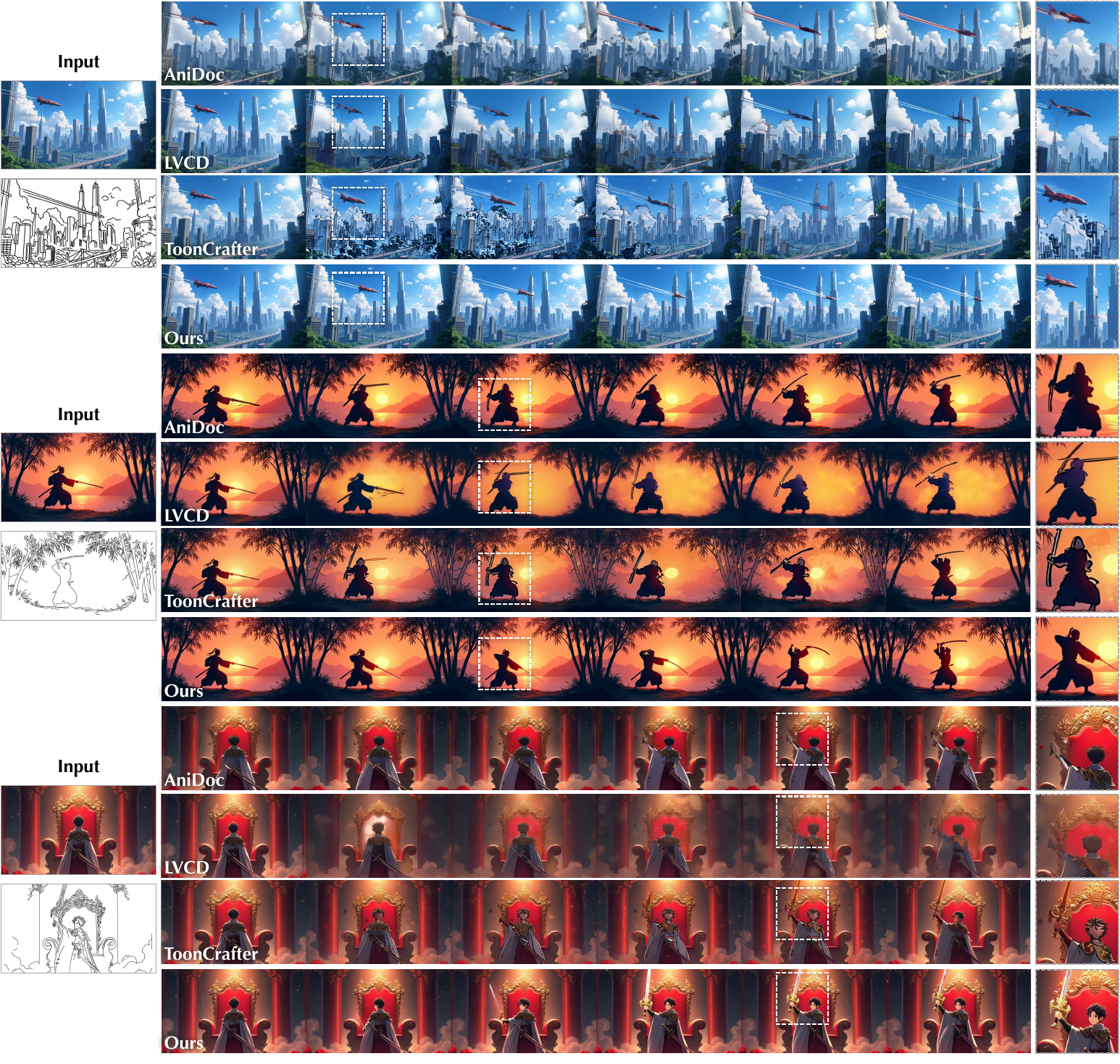}
    \vspace{-4mm}
    \caption{Comparison on the benchmark PKBench, using keyframe sketches drawn by the human artists. Zoom-in patches are shown in the rightmost column. Our method generates high-quality results from real sketch inputs, whereas other methods struggle to maintain visual consistency. Please refer to the supplementary video for additional examples.}
    \label{fig:exp-comp-real}
    \vspace{-4mm}
\end{figure*}

\Cref{fig:exp-comp-real} visualizes the comparison among all methods, with zoom-in patches from a randomly selected region provided in the rightmost column. It is observed that previous methods deviate from the overall style of the first reference frame. Specifically, ToonCrafter generates intermediate frames with prominent bold lines, likely influenced by the bold brush strokes in the human-drawn sketches, revealing its limited robustness to diverse sketch styles. In contrast, our ToonComposer produces video frames with superior visual quality, motion coherence, and style consistency, consistent with the quantitative results.

\begin{figure}
    \centering
    \includegraphics[width=1\linewidth]{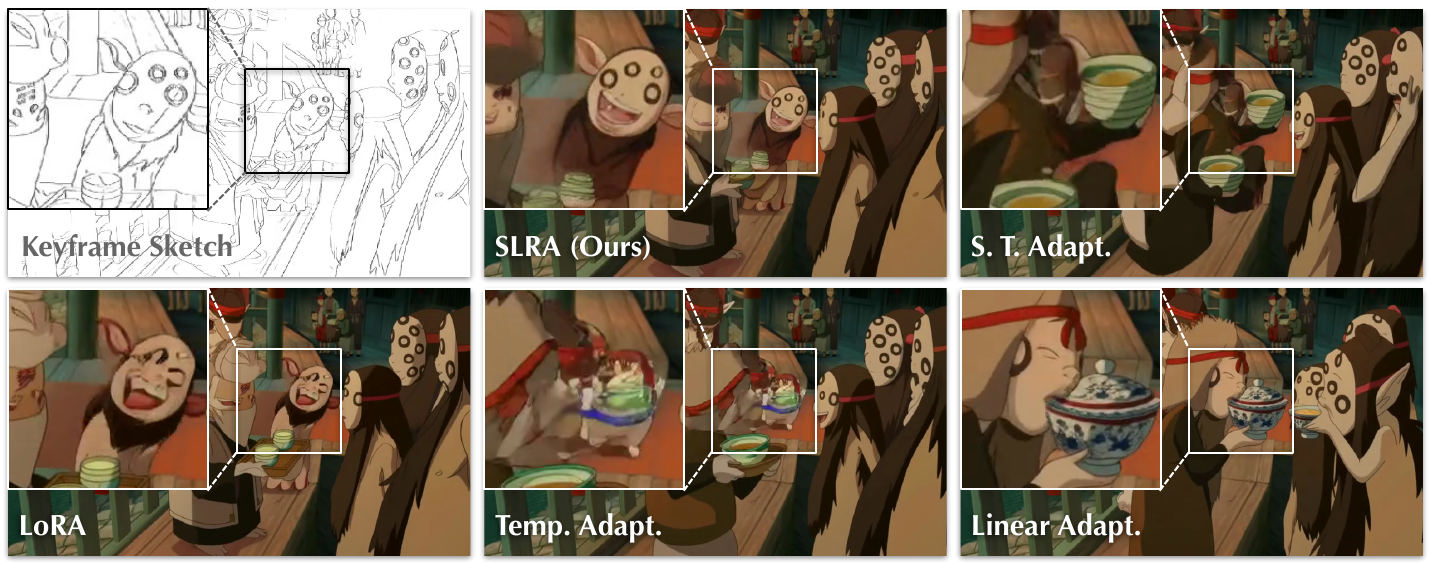}
    \vspace{-6mm}
    \caption{Ablation study of the Spatial Low-Rank Adapter (SLRA) in ToonComposer, comparing cartoon output frames produced using different adaptation methods. SLRA yields higher visual quality and better coherence with the input keyframe sketches compared to alternative approaches.}
    \label{fig:abaltion-slra}
    \vspace{-5mm}
\end{figure}

\begin{figure*}[ht]
    \centering
    \includegraphics[width=1\linewidth]{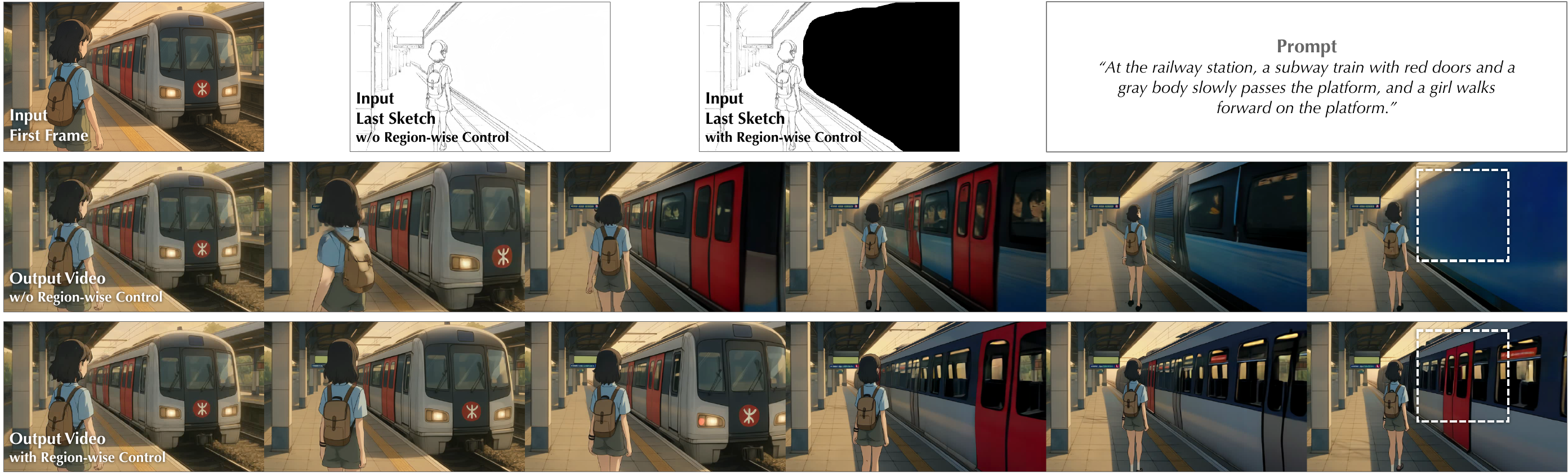}
    \caption{Illustration of region-wise control in ToonComposer. Without region-wise control, blank areas in keyframe sketches are misinterpreted as textureless regions, producing a flat blue train (second row, highlighted with a dashed box). With region-wise control, users can specify areas for context-driven generation without explicit sketches, enabling the model to create plausible and detailed content, such as the dynamic train motion (third row, highlighted with a dashed box).}
    \label{fig:exp-case-region-wise-control}
\end{figure*}

\subsection{Human Evaluation}
To further investigate users' preferences on the generation results, we conducted human evaluations to compare the generation results produced by our method and other baselines. We randomly select 30 samples from the benchmarks and generate cartoon videos for each method using the aforementioned pipeline. 
Our evaluation process involved 47 participants, each of whom was asked to select the video with the best aesthetic quality and motion quality.
The results are shown in~\Cref{tab:human_eval}, where our method achieves the highest win rate on both metrics, significantly exceeding the second best competitor.

\begin{figure*}[t]
    \centering
    \includegraphics[width=1\linewidth]{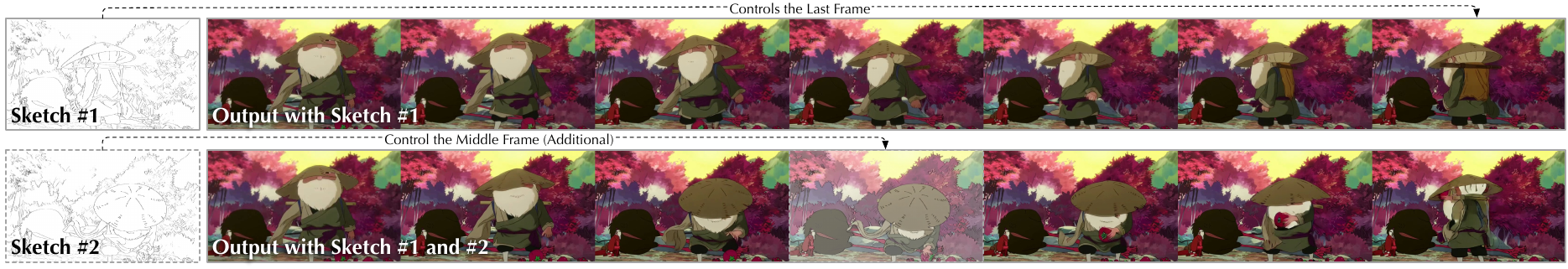}
    \caption{ToonComposer's flexible controllability with varying keyframe sketches. Using only sketch \#1 as the final keyframe and the prompt ``an old man turns back,'' ToonComposer generates a sequence where the old man turns directly (first row). Adding sketch \#2 to control the middle keyframe, while keeping the prompt unchanged, results in a sequence where the old man first picks up a fruit before turning back (second row).}
    \label{fig:exp-multiple-frame}
\end{figure*}

\subsection{Discussion and Analysis}
\label{sec:discussion-analysis}

\paragraph{Ablation on the SLRA}
To assess the importance of spatial adaptation in ToonComposer, we conducted an ablation study on the SLRA, with results detailed in \Cref{fig:method-slra}. We modify SLRA’s internal attention mechanism to explore alternative adaptation behaviors:
a) temporal adaptation (\textit{Temp. Adapt.}), which focuses on temporal dynamics; 
b) spatial-temporal adaptation (\textit{S.T. Adapt.}), which jointly adjusts both; 
c) a degraded linear adapter (\textit{Linear Adapt.}), which removes the attention block entirely. 
d) a baseline using LoRA~\cite{hu2022lora}, which modifies all linear layers (query, key, value, and output) in DiT’s attention modules. This design implicitly alters both spatial and temporal behaviors. To ensure fairness, LoRA’s rank was set to 24 to match the trainable parameter count of SLRA.
All models were trained with identical settings. LPIPS, DISTS, and CLIP image similarity are used for evaluation. %

The results are presented in \Cref{tab:exp-ablation-slra} and \Cref{fig:abaltion-slra}, where SLRA outperforms all variants in both numeric results and visual quality.
Specifically, \textit{ a) Temp. Adapt.} and \textit{ b) S.T. Adapt.} yield higher errors due to insufficient or conflicting spatial adjustments, while \textit{c) Linear Adapt.} lacks the nuanced adaptation required for cartoon aesthetics.
Despite the broader scope of d) LoRA, it underperforms SLRA due to its less targeted adaptation, which disrupts the temporal priors critical for a smooth transition. These findings underscore SLRA’s effectiveness in adapting DiT’s spatial behavior for cartoon-specific features, while preserving the temporal prior intact.

\vspace{-1em}
\paragraph{Use Case of Region-wise Control}
We visualize how region-wise control affects the generated video.
Without region-wise control, leaving a blank area in the keyframe sketch causes the model to interpret it as a textureless region, resulting in flat areas in the generated frames, as illustrated in the second row of \Cref{fig:exp-case-region-wise-control}. 
In contrast, with the region-wise control enabled, users can simply draw an area with brush tools to indicate areas that require generating proper motion according to the context.
As shown in the last row of~\Cref{fig:exp-case-region-wise-control}, our model is able to infer from the input keyframe, the sketch, and the mask given, and automatically generate a plausible movement of the train in the masked area. 
This mechanism significantly improves flexibility, further alleviating manual workload in real scenarios.

\paragraph{Controllability with Increasing Keyframe Sketches}
The sparse sketch injection mechanism of ToonComposer enables flexible control by supporting a variable number of input keyframe sketches, increasing its utility in the cartoon production pipeline. This adaptability allows artists to balance creative control and automation based on the complexity of the desired motion. As shown in \Cref{fig:exp-multiple-frame}, we demonstrate the ability of ToonComposer to generate distinct cartoon sequences from different numbers of input sketches, all conditioned on the same text prompt. Additional examples are available in the supplementary video, which illustrates the versatility of our method in diverse scenarios.

\paragraph{Generalization to 3D Animation}
Despite differences in production pipelines, ToonComposer extends its applicability to 3D-rendered animation by adapting the initial reference frame to a 3D-rendered image. We fine-tuned the model on a compact dataset of 3D animation clips, enabling it to generate high-quality 3D-style sequences in the post-keyframing manner. This adaptability highlights ToonComposer’s versatility and potential for broader animation applications. These 3D animation samples are provided in the supplementary video.

\section{Conclusion}
In this paper, we present ToonComposer, a novel model that streamlines cartoon production by automating tedious tasks of inbetweening and colorization through a unified generative process named \textit{post-keyframing}.
Built on the DiT architecture, ToonComposer leverages sparse keyframe sketches and a single colored reference frame to produce high-quality, stylistically consistent cartoon video sequences.
Experiments show that ToonComposer surpasses existing methods in visual fidelity, motion coherence, and production efficiency. Features such as sparse sketch injection and region-wise control empower artists with precision and flexibility, making ToonComposer a versatile system for cartoon creation. Despite limitations such as computational costs, ToonComposer offers a promising solution to streamline the cartoon production pipeline through generative models.

\section*{Acknowledgments}
We express our sincere gratitude to \textit{B\&T Studio} and \textit{Gudong Animation Studio} for generous permission to use their animation content in our work.

%% file: vars.tex
\providecommand{\newstagename}{post-keyframing\xspace}
\providecommand{\newstagenamecaptial}{Post-Keyframing\xspace}
\providecommand{\datasetname}{PKData\xspace}
\providecommand{\benchmarkname}{PKBench\xspace}